%% file: AICAS2024.tex
\documentclass[conference]{IEEEtran}

\usepackage{graphicx}
\usepackage{cite}
\usepackage{amsmath,amssymb,amsfonts}
\usepackage{algorithmic}
\usepackage{graphicx}
\usepackage{textcomp}
\usepackage{xcolor}
\usepackage{mathtools}
\usepackage{tikz}
\usetikzlibrary{positioning}

\usepackage{standalone}
\usepackage[a4paper, total={184mm,239mm}]{geometry}
\usepackage{multirow} 

\usepackage{etoolbox}
\makeatletter
\newcommand{\linebreakand}{%
  \end{@IEEEauthorhalign}
  \hfill\mbox{}\par
  \mbox{}\hfill\begin{@IEEEauthorhalign}
}

\patchcmd{\@makecaption}
  {\scshape}
  {}
  {}
  {}
\makeatother

\author{\IEEEauthorblockN{Cédric Gernigon}
	\IEEEauthorblockA{Univ. Rennes, Inria, CNRS, IRISA\\ F-35000 Rennes, France\\
Email: cedric.gernigon@inria.fr}
\and
	\IEEEauthorblockN{Silviu-Ioan Filip}
	\IEEEauthorblockA{Univ. Rennes, Inria, CNRS, IRISA\\ F-35000 Rennes, France\\
		Email: silviu.filip@inria.fr}
\and
\IEEEauthorblockN{Olivier Sentieys}
	\IEEEauthorblockA{Univ. Rennes, Inria, CNRS, IRISA\\ F-35000 Rennes, France\\
Email: olivier.sentieys@inria.fr}
\linebreakand
\IEEEauthorblockN{Clément Coggiola}
\IEEEauthorblockA{
Spacecraft techniques, on-board data handling\\
CNES, Toulouse, France \\
Email: clement.coggiola@cnes.fr}
\and
\IEEEauthorblockN{Mickaël Bruno}
\IEEEauthorblockA{
Spacecraft techniques, on-board data handling\\
CNES, Toulouse, France \\
Email: mickael.bruno@cnes.fr}
}

\renewcommand{\paragraph}[1]{\noindent\textbf{#1}\quad}

\begin{document}

\title{AdaQAT: Adaptive Bit-Width Quantization-Aware Training\\}

\author{\IEEEauthorblockN{Cédric Gernigon}
	\IEEEauthorblockA{Univ. Rennes, Inria, CNRS, IRISA\\ F-35000 Rennes, France\\
Email: cedric.gernigon@inria.fr}
\and
\IEEEauthorblockN{Silviu-Ioan Filip}
\IEEEauthorblockA{Univ. Rennes, Inria, CNRS, IRISA\\ F-35000 Rennes, France\\
    Email: silviu.filip@inria.fr}
\and
\IEEEauthorblockN{Olivier Sentieys}
	\IEEEauthorblockA{Univ. Rennes, Inria, CNRS, IRISA\\ F-35000 Rennes, France\\
Email: olivier.sentieys@inria.fr}
\linebreakand
\IEEEauthorblockN{Clément Coggiola}
\IEEEauthorblockA{
Spacecraft techniques, on-board data handling\\
CNES, Toulouse, France \\
Email: clement.coggiola@cnes.fr}
\and
\IEEEauthorblockN{Mickaël Bruno}
\IEEEauthorblockA{
Spacecraft techniques, on-board data handling\\
CNES, Toulouse, France \\
Email: mickael.bruno@cnes.fr}
}

\maketitle

\begin{abstract}
Large-scale deep neural networks (DNNs) have achieved remarkable success in many application scenarios. However, high computational complexity and energy costs of modern DNNs make their deployment on edge devices challenging. Model quantization is a common approach to deal with deployment constraints, but searching for optimized bit-widths can be challenging. In this work, we present \textbf{Ada}ptive Bit-Width \textbf{Q}uantization \textbf{A}ware \textbf{T}raining (\textbf{AdaQAT}), a learning-based method that automatically optimizes weight and activation signal bit-widths during training for more efficient DNN inference. We use relaxed real-valued bit-widths that are updated using a gradient descent rule, but are otherwise discretized for all quantization operations. The result is a simple and flexible QAT approach for mixed-precision uniform quantization problems. Compared to other methods that are generally designed to be run on a pretrained network, AdaQAT works well in both training from scratch and fine-tuning scenarios.
Initial results on the CIFAR-10 and ImageNet datasets using ResNet20 and ResNet18 models, respectively, indicate that our method is competitive with other state-of-the-art mixed-precision quantization approaches.
\end{abstract}

\begin{IEEEkeywords}
Neural Network Compression, Quantization Aware Training, Adaptive Bit-Width Optimization
\end{IEEEkeywords}

\section{Introduction}
\label{sec:intro}
Deep Neural Networks (DNN) have achieved remarkable results in recent years in a wide range of domains. While many inference computations are done in the cloud, it is increasingly desirable to deploy trained DNNs to edge devices, such as mobile phones and wearable devices, due to privacy, security, and latency concerns or limitations in communication bandwidth.
However, modern DNNs contain at least millions of parameters and require billions of arithmetic operations. Memory and computational costs make deployment on embedded devices difficult, if not infeasible in many cases.
To mitigate these issues, various compression techniques have been proposed, such as pruning~\cite{han2015deep}, weight sharing~\cite{dupuis2020automatic}, knowledge distillation~\cite{tung2019similarity}, and quantization~\cite{hubara2017quantized}.
With the emergence of hardware platforms offering better support for low (\emph{e.g.}~recent Nvidia GPUs and Google TPUs) and custom precision (\emph{e.g.}~FPGA and ASIC solutions) compute, quantization is at the forefront of methods used to increase the efficiency of DNN model inference. 

Many models can be uniformly quantized to $8$ bits~\cite{zhoudorefa-net2018} and in some cases to even binary~\cite{courbariaux_binarized_nodate} or ternary~\cite{li_ternary_2016} representations. Various methods~\cite{wang_haq_2019, huang2022sdq, yang2021fracbits, dong2019hawq} push the compression limit even further, using different bit-widths at the (sub)layer level. However, it is challenging to find efficient mixed-precision configurations that compress a model with minimal impact on accuracy. There are three main families of methods that attempt to optimize bit-width allocation for model compression: \emph{search}, \emph{metric} and \emph{optimization}-based. Search-based methods iteratively explore the bit-width assignment space and are generally costly to use. Metric-based approaches are much faster, but tend to give sub-optimal results. Optimization methods offer good performance at a reasonable cost, but most results of this type tend to suffer from instabilities during the optimization process (cf.~\cite{huang2022sdq}), especially if starting far (\emph{e.g.}~training from scratch) from an optimized configuration.

In the following, we introduce \textbf{Ada}ptive Bit-Width \textbf{Q}uanti{-}zation \textbf{A}ware \textbf{T}raining (\textbf{AdaQAT}), an optimization-based method for mixed-precision uniform quantization of both weights and activations. Its defining characteristic is the use of relaxed fractional bit-widths that are updated using a gradient descent rule, but are otherwise discretized for all operations (in forward and backward passes). Compared to previous approaches, our initial tests show that AdaQAT is able to produce efficient quantized DNNs that are comparable to the state of the art, in both training from scratch and fine-tuning settings.

\section{Related Work}
Depending on when quantization is performed (after or during training), there are two main families of methods used in practice. The first, Post-Training Quantization (PTQ), is fast, but can lead to non-negligible loss in accuracy for very small formats~\cite{banner_post_nodate}. Quantization Aware Training (QAT), while slower, generally leads to better results and should be preferred for extreme quantization problems.
 
\subsection{Quantization-Aware Training}

QAT methods stem from pioneering work on binary neural networks~\cite{courbariaux_binaryconnect_2015,courbariaux_binarized_nodate}. At its core, a QAT method consists of using a quantized version of the network during training in both forward and backward passes, while performing updates on full-precision copies of the network parameters. These full precision parameters are then quantized to be used in the next iteration. A crucial aspect is how to perform backpropagation through quantized variables (parameters and activations). In the binary case, this was done using a so-called Straight-Through Estimator (STE)~\cite{bengio_estimating_2013} and this approach was later~\cite{zhoudorefa-net2018} extended to cover larger bit-widths, while also applying quantization to gradient signals.

To further improve the accuracy of quantized DNNs, the STE idea can also be used to learn the parameters of uniform quantizers, such as scaling factors and bias terms for weight quantization~\cite{esser2019learned,bhalgat2020lsq+}, and in the case of ReLU-based activations, clipping parameters~\cite{choipact2018}.

\subsection{Bit-Width Search Strategies}
Finding bit-width allocations that improve inference efficiency has been addressed using various approaches.

Search-based methods like HAQ~\cite{wang_haq_2019} rely on reinforcement learning with hardware (latency \& energy) feedback in the agent, whereas neural architecture search work like DNAS~\cite{wu2018mixed} uses gradient-based information. The major downside in using them is that they require significant time and computational resources.

Much faster results can be obtained using metric-based methods. For example, HAWQ~\cite{dong2019hawq} uses Hessian spectrum information at each layer to assign precisions. Methods like~\cite{yao2021hawq,ma2021ompq} rely on linear programming models, while~\cite{liu2021sharpness} encourages quantization that leads to reduced sharpness in the task loss function. A potential downside of these methods might be the fact that they can lead to sub-optimal results compared to other approaches (cf.~\cite{huang2022sdq}).

Optimization-based approaches formulate the bit-width assignment as an optimization problem, with the main challenge being how to handle the fact that the loss is non-differentiable w.r.t.~the bit-widths. Methods like FracBits~\cite{yang2021fracbits} and BitPruning~\cite{nikolic2020bitpruning} use fractional bit-widths and linear interpolation during the forward path, whereas SDQ~\cite{huang2022sdq} is based on stochastic quantization, but seems limited to weight quantization. These methods work well in fine-tuning scenarios, but are unstable or do not work when training from scratch.

AdaQAT falls into this third category. It is an optimization-based mixed-precision QAT method that shows good flexibility when compared to other approaches in the same vein.

\section{Method}

We start by presenting the necessary background on DNN quantization before to describe in detail the proposed method.

\subsection{Quantization background}
We adopt the DoReFa~\cite{zhoudorefa-net2018} scheme for weight quantization and PACT~\cite{choipact2018} for activation quantization with the improvements suggested in SAT~\cite{jin2019towards}. The same quantization function is applied to both weight and activation:

\begin{equation}
   \texttt{q}(x) = \dfrac{1}{s}\left\lfloor x s\right\rceil,
   \label{eq:quant}
\end{equation}
where \(x \in \left[0, 1\right]\), \(\left\lfloor \cdot \right\rceil\) indicates rounding to the nearest integer, $s=2^k-1$ is the scaling factor and $k$ is the quantization bit-width.

The weight tensors are first brought into $[0,1]$ using the transformation $f(\mathbf{w})=\dfrac{\tanh(\mathbf{w})}{2\max(|\tanh(\mathbf{w})|)}$ and then rescaled and shifted to $[-1,1]$. Backpropagation through \eqref{eq:quant} is done using STE, leading to the following rule for $\mathbf{w}$:
\begin{align*}
    \textbf{Forward: } & \mathbf{w}_q = 2\;\texttt{q}\left(f(\mathbf{w})+\dfrac{1}{2}\right) - 1\\
    \textbf{Backward: } & \dfrac{\partial \mathcal{L}}{\partial \mathbf{w}} = \dfrac{\partial \mathcal{L}}{\partial \mathbf{w}_q}\dfrac{\partial \mathbf{w}_q}{\partial \mathbf{w}}
\end{align*}
where $\mathbf{w}$ is an unquantized weight tensor, $\mathcal{L}$ is the loss function, and $\mathbf{w}_q$ is the quantized version of $\mathbf{w}$.

PACT~\cite{choipact2018} proposes to learn the upper bound of a ReLU activation function in order to compute an appropriate scaling factor $s$. The vanilla ReLU is thus replaced with:
\begin{equation*}
    \text{PACT}(x) = 
\left \{
    \begin{array}{ll}
        0 & \mbox{if } x < 0 \\
        \alpha & \mbox{if } x > \alpha  \\
        x & \mbox{otherwise}
    \end{array}
\right.
\end{equation*}

The scaling factor in~\eqref{eq:quant} is now $s=(2^k-1)/\alpha$. The complete activation quantization procedure is:

\begin{align*}
    \textbf{Forward: } & \mathbf{y}_q = \texttt{q}\left(\mathbf{y}\right)\\
    \textbf{Backward: } & \dfrac{\partial \mathcal{L}}{\partial \mathbf{y}} = \dfrac{\partial \mathcal{L}}{\partial \mathbf{y}_q}\mathbb{I}_{\mathbf{x}\leqslant \alpha} \text{ and } \dfrac{\partial \mathbf{y}_q}{\partial \alpha} = \dfrac{\partial \mathbf{y}}{\partial \alpha}\mathbb{I}_{\mathbf{x}\leqslant \alpha}\\
\end{align*}
where $\mathbf{y}$ is an unquantized activation, $\mathcal{L}$ is the loss function, $\mathbf{y}_q$ is the quantized activation, and $\mathbb{I}_{\mathcal{C}(\mathbf{x})}$ is an indicator function that returns $1$ if $\mathbf{x}$ satisfies condition $\mathcal{C}$ and $0$ otherwise.

\subsection{Objective Function}\label{sec:objective}
In order to learn the bit-widths of the uniform quantizers for both weights and activations, we use two real-valued variables $N_\mathbf{w}$ and $N_\mathbf{a}$, respectively. The actual integer bit-widths of the quantized network are $\lceil N_\mathbf{w}\rceil$ and $\lceil N_\mathbf{a}\rceil$.

We model the loss function to minimize that takes into account the cost of a particular bit-width configuration as:
\begin{equation}\label{eq:total_loss}
    \mathcal{L}_\text{Total}=\mathcal{L}_\text{Task}\left(\left\lceil N_\mathbf{w}\right\rceil, \left\lceil N_\mathbf{a}\right\rceil\right)+\lambda\mathcal{L}_\text{Hard}\left(\left\lceil N_\mathbf{w} \right\rceil, \left\lceil N_\mathbf{a}\right\rceil\right)
\end{equation}
where $\lambda>0$ is a balancing hyper-parameter between the task and hardware losses.

FracBits~\cite{yang2021fracbits} has reviewed various methods used to model the hardware cost of arithmetic precision choices for weights and activations. They argue in favor of memory size if only targeting weight quantization, and BitOPs (see~\cite[eqs.~(4) and (5)]{yang2021fracbits}) for joint weight and activation quantization.
For a convolutional filter $f$, the BitOPs metric corresponds to 
\[
\text{BitOPs}(f) = \left\lceil N_\mathbf{w}\right\rceil \left\lceil N_\mathbf{a}\right\rceil |f| w_f h_f /s^2_f,
\]
where $|f|$ denotes the cardinality of the filter, $w_f$, $h_f$, $s_f$ are the spatial width, height, and stride of the filter, respectively.

In our particular case, since we are using one bit-width per weights and one per activations, the overall BitOPs hardware cost will be linear in $\lceil N_\mathbf{w}\rceil\lceil N_\mathbf{a}\rceil$, namely

\begin{equation*}
    \mathcal{L}_\text{Hard}\left(\left\lceil N_\mathbf{w} \right\rceil, \left\lceil N_\mathbf{a}\right\rceil\right)=\left\lceil N_\mathbf{w}\right\rceil\left\lceil N_\mathbf{a}\right\rceil.
\end{equation*}

\subsection{Bit-Width Gradients \& Parameter Updates}
Since the task loss is not directly differentiable with respect to the bit-width parameters, we use finite difference approximations as follows:
\begin{equation*}
\begin{split}
    \dfrac{\partial \mathcal{L}_\text{Task}}{\partial N_\mathbf{w}} & \approx  \mathcal{L}_\text{Task}(\left\lceil N_\mathbf{w}\right\rceil, \left\lceil N_\mathbf{a}\right\rceil)-\mathcal{L}_\text{Task}(\left\lfloor N_\mathbf{w}\right\rfloor,\left\lceil N_\mathbf{a}\right\rceil) \\
    \dfrac{\partial \mathcal{L}_\text{Task}}{\partial N_\mathbf{a}} & \approx  \mathcal{L}_\text{Task}(\left\lceil N_\mathbf{w}\right\rceil, \left\lceil N_\mathbf{a}\right\rceil)-\mathcal{L}_\text{Task}(\left\lceil N_\mathbf{w}\right\rceil, \left\lfloor N_\mathbf{a}\right\rfloor)
\end{split}
\end{equation*}

\begin{figure*}[ht]
    \centering
    \includegraphics[width=\textwidth]{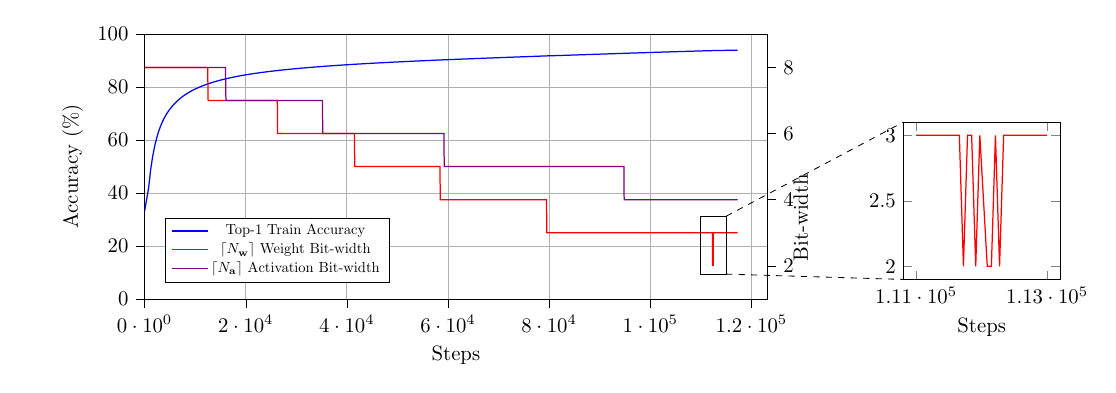}
    \caption{Example of applying our approach with a ResNet20 network on the CIFAR-10 dataset. It showcases the evolution of the train accuracy with respect to updating the bit-width parameters $\left\lceil N_\mathbf{w}\right\rceil$ and $\left\lceil N_\mathbf{a}\right\rceil$ and how an oscillatory pattern can form (here, for the weight bit-width $\left\lceil N_\mathbf{w}\right\rceil$). When oscillations appear, we fix the value of the corresponding bit-width to the largest of the two oscillation points for the rest of the QAT process, considering that it has converged.}
    \label{img:results}
\end{figure*}

The gradient of the total loss w.r.t.~the bit-widths is then approximated as:
\begin{equation}\label{eq:bitwidth_grad}
\begin{split}
    \dfrac{\partial \mathcal{L}_\text{Total}}{\partial  N_\mathbf{x}} & \approx \dfrac{\partial \mathcal{L}_\text{Task}}{\partial N_\mathbf{x}} + \lambda\dfrac{\partial \mathcal{L}_\text{Hard}}{\partial \left\lceil N_\mathbf{x}\right\rceil}
\end{split}
\end{equation}
which are then used to update the fractional bit-width parameters. The gradient descent rule that does this takes the form
\begin{equation}\label{eq:bitwidth_update}
    N_\mathbf{x}^+ = N_\mathbf{x} - \eta_\mathbf{x}\dfrac{\partial \mathcal{L}_\text{Total}}{\partial N_\mathbf{x}},
\end{equation}
with $\mathbf{x}\in \left\lbrace\mathbf{w},\mathbf{a}\right\rbrace$, $N_\mathbf{x}^+$ the new bit-width values at the next iteration, and $\eta_\mathbf{x}>0$ corresponding learning rates.

The rest of the network and quantizer parameters are updated using the SGD-like or accelerated algorithms that train the network normally, with their own hyperparameters.

We have noticed that too rapid changes in the values of the learned bit-widths tend to degrade accuracy considerably, slowing down the optimization process. To avoid this, the learning rates need to be reasonably small. Unless otherwise stated, default values of $\eta_\mathbf{w}=0.001$ and $\eta_\mathbf{a}=0.0005$ are considered in our testing. A smaller learning rate $\eta_\mathbf{a}$ is picked since it appears that the progressive quantization of activations is more sensitive to changes in $N_\mathbf{a}$ than weight quantization is to changes in $N_\mathbf{w}$.

When $N_\mathbf{w}$ and $N_\mathbf{a}$ reach their optimized values, continuing to decrease them will lead to a (steep) increase of the task loss $\mathcal{L}_\text{Task}$ and consequently of $\mathcal{L}_\text{Total}$. This means that their gradient estimates~\eqref{eq:bitwidth_grad} will become negative and~\eqref{eq:bitwidth_update} will start increasing $N_\mathbf{w}$ and $N_\mathbf{a}$. An oscillatory pattern forms. For an example, see Figure~\ref{img:results}. When this happens, we monitor the number of oscillations and as soon as it passes a certain threshold (which we empirically set to $10$) we fix the bit-widths to $\left\lceil N_\mathbf{w}\right\rceil$ and $\left\lceil N_\mathbf{a}\right\rceil$, respectively, and continue the rest of the quantization process in standard QAT fashion.




\section{Experiments}

To evaluate the effectiveness of AdaQAT, we conduct several mixed-precision quantization experiments on the CIFAR-10~\cite{krizhevsky2010cifar} and the ImageNet~\cite{deng2009imagenet} datasets and compare the results with those obtained with other mixed-precision quantization methods from the state-of-the-art.

\subsection{Experimental Setup}
\paragraph{Datasets}
We use the CIFAR-10 and ImageNet datasets for our experiments. We only perform basic data augmentation on the training dataset~\cite{lee2015deeply}, which includes (in PyTorch parlance) \textit{RandomResizedCrop} and \textit{RandomHorizontalFlip} during training, and a single-crop operation during evaluation for ImageNet.

\paragraph{Networks} We use a ResNet20~\cite{he2016deep} model on CIFAR-10 and a ResNet18~\cite{he2016deep} one on ImageNet. Following the practice adopted by prior work regarding greater sensitivity to quantization at the input and output of a network (see for instance~\cite{yang2021fracbits}), we fix the bit-width to $8$ bits in the first and last layers.

\paragraph{Training Settings} We use an SGD optimizer with a batch size of $256$, weight decay set to $10^{-4}$, and momentum to $0.9$. In the training from scratch scenario weights are initialized using the Kaiming method~\cite{he2015delving}. We use a cosine annealing learning rate scheduler with initial learning rate set to $0.1$ for the from scratch scenario and $0.01$ for the fine-tuning scenario. Training is run for $150$ epochs in the fine-tuning scenario and $300$ epochs when starting from scratch. We use PyTorch 1.13 for all experiments. The ImageNet tests are run on a cluster of $8$ NVIDIA V100 GPUs, whereas the CIFAR-10 ones use a single GPU configuration.

\subsection{Comparison with State-of-the-Art Methods}
Table~\ref{tab:SoA} shows the results of applying AdaQAT on CIFAR-10 compared to other methods from the literature. The first line shows the floating-point baseline result, whereas the second group of lines showcases \emph{static} methods, where activations are not quantized and weights are quantized uniformly to $2$ bits. The third group of lines corresponds to mixed-precision methods where the weight bit-width is learned and the activations are not quantized. AdaQAT with learned weight bit-width ($2$ bits) and unquantized activations is on par with the best of these, both when starting from a pretrained full-precision model as well as from scratch. We should nevertheless note that the FracBits results were obtained without fine-tuning its hyperparameters as much as possible.

The last two groups of lines in Table~\ref{tab:SoA} illustrate the behaviour of our method when the activations are also quantized. The accuracy results are still competitive, either when starting from a pretrained model or from scratch. Even though the WCR metric is not as good as that of SDQ, it is more than compensated by the reduction in activation bit-width. It directly impacts how much memory (the BitOps column) gets transferred from one layer of the network to the next, going from $2.61$ down to $0.51$, more than a $5\times$ improvement.

Table~\ref{tab:ImageNet} shows similar results on ImageNet compression. The quality of the obtained quantization is comparable to other methods from the state of the art. SDQ uses knowledge distillation with a ResNet-101 model as teacher, coupled with color jitter data augmentation, leading to better accuracy.

\begin{table}[tb]
\centering
    \caption{Comparison with state-of-the-art quantization methods (ResNet20 on CIFAR-10). Bit-width (W/A) denotes the average bit-width for weights and activation signals, whereas WCR represents the weight compression rate w.r.t. baseline. BitOPs denotes the bit operations metric (see Sec.~\ref{sec:objective}). The $4$-bit activation result is learned using our method ($\lambda=0.15$), whereas the activation bit-widths are fixed in the $8$-bit and $32$-bit settings, with only the weight bit-widths being learned.}
    \label{tab:SoA}
    \begin{tabular}{  c  c  c  c  c c }
    \hline
    \multirow{2}{*}{Method} & Bit-width & {top-1} & {$\Delta_\text{acc}$} & \multirow{2}{*}{WCR} & BitOPs \\ 
    & (W/A) & (\%) & (\%) & & (Gb)\\ 
    \hline
    baseline & 32/32 & 92.4 & - & - & 41.7 \\
    \hline
    DoReFa~\cite{zhoudorefa-net2018} & 2/32 & 88.2 & -4.2 & 16$\times$ & 2.7 \\
    PACT~\cite{choipact2018} & 2/32 & 89.7 & -2.7 & 16$\times$ & 2.7\\
    LQ-Net~\cite{zhang2018lq} & 3/3 & 91.6 & -0.5 & 10.7$\times$ & 0.39\\
    \hline
    FracBits~\cite{yang2021fracbits} & 2.00/32 & 89.6 & -2.8 & 16$\times$ & - \\
    TTQ~\cite{jain2020trained} & 2.00/32 & 91.1 & -1.2 & 16$\times$ & - \\
    SDQ~\cite{huang2022sdq} & 1.93/32 & 92.1 & -0.3 & 16.6$\times$ & - \\
    HAWQ-V1~\cite{dong2019hawq} & 3.89/4 & 92.2 & -0.2 & 8.2$\times$ & 0.67 \\ 
    \hline
    \textbf{\multirow{3}{*}{{\shortstack{Ours\\(fine-tuning)}}}} & 2/32 & 92.0 & -0.4 & 16$\times$ & 2.7 \\
    & 3/8 & 92.1 & -0.3 & 10.7$\times$ & 0.99 \\
    & \textbf{3/4} & \textbf{92.2} & \textbf{-0.2} & \textbf{10.7$\times$} & \textbf{0.51} \\
    \hline
    \multirow{3}{*}{\textbf{\shortstack{Ours\\(from scratch)}}} & 2/32 & 91.8 & -0.6 & 16$\times$ & 2.7\\
    & 3/8 & 91.8 & -0.6 & 10.7$\times$  & 0.99 \\
    & \textbf{3/4} & \textbf{92.1} & \textbf{-0.3} & \textbf{10.7$\times$} & \textbf{0.51}\\
    \hline
    \end{tabular}
\end{table}

\begin{table}[tb]
    \caption{Comparison with state-of-the-art quantization methods on the ImageNet dataset with ResNet18 in a fine-tuning setting. We set $\lambda$ in our approach to $0.15$.}
    \begin{center}
    \label{tab:ImageNet}
    \begin{tabular}{ c  c  c  c  c  c }
    \hline
    \multirow{2}{*}{Method} & Bit-width & \multicolumn{2}{c}{Accuracy(\%)} & \multirow{2}{*}{WCR} & BitOPs \\ 
    & (W/A) & top-1 & FP top-1 & & (Gb)\\ 
    \hline
    DoReFa~\cite{zhoudorefa-net2018} & 4/4 & 68.1 & 70.5 & $8\times$ & 35.2 \\
    PACT~\cite{choipact2018} & 4/4 & 69.2 & 70.5 & $8\times$ & 35.2 \\
    LQ-Net~\cite{zhang2018lq} & 4/4 & 69.3 & 70.3 & $8\times$ & 35.2 \\
    \hline
    FracBits~\cite{yang2021fracbits} & 4.00/4.00 & 70.6 & 70.2 & $8\times$ & 34.7 \\
    SDQ~\cite{huang2022sdq} & 3.85/4 & 71.7 & 70.5 & $8.9\times$ & 33.4 \\
    HAWQ-V3~\cite{yao2021hawq} & 4.8/7.5 & 70.4 & 71.5 & $6.7\times$ & 72.0 \\
    \hline
    \multirow{1}{*}{\textbf{\shortstack{Ours}}} & \textbf{4/4} & \textbf{70.3} & \textbf{70.5} & \textbf{8$\times$} & \textbf{35.2} \\
    \hline
    \end{tabular}
    \end{center}
\end{table}

\begin{table}[ht]
\begin{center}
    \caption{Evolution of AdaQAT mixed-precision quantization results on CIFAR-10 with respect to $\lambda$.}
    \label{tab:balancing}
    \begin{tabular}{  c  c  c  c }
    \hline
    $\lambda$ & W & A & top-1\\ \hline
    0.2 & 2 & 4 & 91.7\\
    0.15 & 3 & 4 & 92.1\\
    0.1 & 4 & 5 & 92.3\\
    \hline
    \end{tabular}
\end{center}
\end{table}

\subsection{Balancing Parameter Impact}

The hyperparameter $\lambda$ dictates how the task loss $\mathcal{L}_\text{Task}$ and the hardware complexity $\mathcal{L}_\text{Hard}$ are balanced out in the total loss $\mathcal{L}_\text{Total}$ (see eq.~\eqref{eq:total_loss}). It controls how much accuracy loss is allowed in the final DNN model compared to the W/A quantization levels. As can be seen in Table~\ref{tab:balancing}, a larger $\lambda$ leads to more compression, but less accurate test results as well. Its value should be chosen carefully on a model-by-model basis, taking into account the application constraints (\emph{i.e.,}~how much accuracy degradation is allowed versus a certain level of attainable compression).

\section{Conclusion \& Future Work}
We have introduced AdaQAT, an optimization-based method for mixed-precision quantization. Compared to previous approaches that are generally intended to be used in a fine-tuning setting, in early tests AdaQAT seems to be more flexible, being capable of operating in both fine-tuning and training from scratch scenarios, producing results that are on par with state-of-the-art mixed-precision quantization approaches on CIFAR10 with a ResNet20 network. It also performs well in mixed-precision fine-tuning of ResNet18 on ImageNet.

As future work, we will evaluate AdaQAT on other network types that are more sensitive to quantization (\emph{e.g.}~the MobileNet family of models). Currently, bit-width assignment is done on a per-network basis. Our goal is to generalize the approach to cover a much larger design space. One direction is to look at finer levels of mixed-precision quantization granularity, such as per-layer and per-channel. We also intend to explore finer hardware complexity and energy consumption metrics, tailored for a specific target architecture (\emph{e.g.}~FPGAs), in the $\mathcal{L}_\text{Hard}$ term.  

\newpage
\bibliographystyle{./IEEEtran}
\input{references.bbl}

\end{document}

%% file: references.bbl

%% file: AICAS2024.bbl
\begin{thebibliography}{10}
\providecommand{\url}[1]{#1}
\csname url@samestyle\endcsname
\providecommand{\newblock}{\relax}
\providecommand{\bibinfo}[2]{#2}
\providecommand{\BIBentrySTDinterwordspacing}{\spaceskip=0pt\relax}
\providecommand{\BIBentryALTinterwordstretchfactor}{4}
\providecommand{\BIBentryALTinterwordspacing}{\spaceskip=\fontdimen2\font plus
\BIBentryALTinterwordstretchfactor\fontdimen3\font minus
  \fontdimen4\font\relax}
\providecommand{\BIBforeignlanguage}[2]{{%
\expandafter\ifx\csname l@#1\endcsname\relax
\typeout{** WARNING: IEEEtran.bst: No hyphenation pattern has been}%
\typeout{** loaded for the language `#1'. Using the pattern for}%
\typeout{** the default language instead.}%
\else
\language=\csname l@#1\endcsname
\fi
#2}}
\providecommand{\BIBdecl}{\relax}
\BIBdecl

\bibitem{han2015deep}
S.~Han, H.~Mao, and W.~J. Dally, ``{Deep Compression: Compressing Deep Neural
  Networks with Pruning, Trained Quantization and Huffman Coding},''
  \emph{arXiv:1510.00149}, 2015.

\bibitem{dupuis2020automatic}
E.~Dupuis, D.~Novo, I.~O’Connor, and A.~Bosio, ``{On the Automatic
  Exploration of Weight Sharing for Deep Neural Network Compression},'' in
  \emph{2020 Design, Automation \& Test in Europe Conference \& Exhibition
  (DATE)}.\hskip 1em plus 0.5em minus 0.4em\relax IEEE, 2020, pp. 1319--1322.

\bibitem{tung2019similarity}
F.~Tung and G.~Mori, ``{Similarity-Preserving Knowledge Distillation},''
  \emph{IEEE/CVF Int. Conf. on Computer Vision}, pp. 1365--1374, 2019.

\bibitem{hubara2017quantized}
I.~Hubara, M.~Courbariaux, D.~Soudry, R.~El-Yaniv, and Y.~Bengio, ``{Quantized
  Neural Networks: Training Neural Networks with Low Precision Weights and
  Activations},'' \emph{The Journal of Machine Learning Research}, vol.~18,
  no.~1, pp. 6869--6898, 2017.

\bibitem{zhoudorefa-net2018}
S.~Zhou, Y.~Wu, Z.~Ni, X.~Zhou, H.~Wen, and Y.~Zou, ``{DoReFa-Net: Training Low
  Bitwidth Convolutional Neural Networks with Low Bitwidth Gradients},''
  \emph{arXiv:1606.06160}, 2016.

\bibitem{courbariaux_binarized_nodate}
M.~Courbariaux, I.~Hubara, D.~Soudry, R.~El-Yaniv, and Y.~Bengio, ``{Binarized
  Neural Networks: Training Deep Neural Networks with Weights and Activations
  Constrained to $+1$ or $-1$},'' \emph{arXiv:1602.02830}, 2016.

\bibitem{li_ternary_2016}
F.~Li, B.~Zhang, and B.~Liu, ``Ternary {Weight} {Networks},''
  \emph{arXiv:1605.04711}, 2016.

\bibitem{wang_haq_2019}
K.~Wang, Z.~Liu, Y.~Lin, J.~Lin, and S.~Han, ``{HAQ}: {Hardware}-{Aware}
  {Automated} {Quantization} {With} {Mixed} {Precision},'' \emph{IEEE/CVF Int.
  Conf. on Computer Vision}, pp. 8604--8612, 2019.

\bibitem{huang2022sdq}
X.~Huang, Z.~Shen, S.~Li, Z.~Liu, H.~Xianghong, J.~Wicaksana, E.~Xing, and
  K.-T. Heng, ``{SDQ: Stochastic Differentiable Quantization with Mixed
  Precision},'' \emph{Int. Conf. on Machine Learning}, pp. 9295--9309, 2022.

\bibitem{yang2021fracbits}
L.~Yang and Q.~Jin, ``{FracBits: Mixed Precision Quantization via Fractional
  Bit-Widths},'' \emph{AAAI Conf. on Artificial Intelligence}, pp.
  10\,612--10\,620, 2021.

\bibitem{dong2019hawq}
Z.~Dong, Z.~Yao, A.~Gholami, M.~Mahoney, and K.~Keutzer, ``{HAWQ: Hessian Aware
  Quantization of Neural Networks with Mixed-Precision},'' \emph{IEEE/CVF Int.
  Conf. on Computer Vision}, pp. 293--302, 2019.

\bibitem{banner_post_nodate}
R.~Banner, Y.~Nahshan, and D.~Soudry, ``Post training 4-bit quantization of
  convolutional networks for rapid-deployment,'' \emph{Neural Information
  Processing Systems}, vol.~32, 2019.

\bibitem{courbariaux_binaryconnect_2015}
M.~Courbariaux, Y.~Bengio, and J.-P. David, ``{BinaryConnect}: {Training}
  {Deep} {Neural} {Networks} with binary weights during propagations,''
  \emph{{Neural} {Information} {Processing} {Systems}}, vol.~2, pp. 3123--3131,
  2015.

\bibitem{bengio_estimating_2013}
Y.~Bengio, N.~L{\'e}onard, and A.~Courville, ``{Estimating or Propagating
  Gradients Through Stochastic Neurons for Conditional Computation},''
  \emph{arXiv:1308.3432}, 2013.

\bibitem{esser2019learned}
S.~Esser, J.~McKinstry, D.~Bablani, R.~Appuswamy, and D.~Modha, ``{Learned Step
  Size Quantization},'' \emph{arXiv:1902.08153}, 2019.

\bibitem{bhalgat2020lsq+}
Y.~Bhalgat, J.~Lee, M.~Nagel, T.~Blankevoort, and N.~Kwak, ``{LSQ+: Improving
  Low-Bit Quantization Through Learnable Offsets and Better Initialization},''
  \emph{IEEE/CVF Int. Conf. on Computer Vision}, pp. 696--697, 2020.

\bibitem{choipact2018}
J.~Choi, Z.~Wang, S.~Venkataramani, P.~Chuang, V.~Srinivasan, and
  K.~Gopalakrishnan, ``{PACT: Parameterized Clipping Activation for Quantized
  Neural Networks},'' \emph{arXiv:1805.06085}, 2018.

\bibitem{wu2018mixed}
B.~Wu, Y.~Wang, P.~Zhang, Y.~Tian, P.~Vajda, and K.~Keutzer, ``{Mixed Precision
  Quantization of ConvNets via Differentiable Neural Architecture Search},''
  \emph{arXiv:1812.00090}, 2018.

\bibitem{yao2021hawq}
Z.~Yao, Z.~Dong, Z.~Zheng, A.~Gholami, J.~Yu, E.~Tan, L.~Wang, Q.~Huang,
  Y.~Wang, and M.~Mahoney, ``{HAWQ-V3: Dyadic Neural Network Quantization},''
  \emph{Int. Conf. on Machine Learning}, pp. 11\,875--11\,886, 2021.

\bibitem{ma2021ompq}
Y.~Ma, T.~Jin, X.~Zheng, Y.~Wang, H.~Li, Y.~Wu, G.~Jiang, W.~Zhang, and R.~Ji,
  ``{OMPQ: Orthogonal Mixed Precision Quantization},'' \emph{arXiv:2109.07865},
  2021.

\bibitem{liu2021sharpness}
J.~Liu, J.~Cai, and B.~Zhuang, ``{Sharpness-Aware Quantization for Deep Neural
  Networks},'' \emph{arXiv:2111.12273}, 2021.

\bibitem{nikolic2020bitpruning}
M.~Nikoli{\'c}, G.~Hacene, C.~Bannon, A.~Lascorz, M.~Courbariaux, Y.~Bengio,
  V.~Gripon, and A.~Moshovos, ``{BitPruning: Learning Bitlengths for Aggressive
  and Accurate Quantization},'' \emph{arXiv:2002.03090}, 2020.

\bibitem{jin2019towards}
Q.~Jin, L.~Yang, and Z.~Liao, ``Towards efficient training for neural network
  quantization,'' \emph{arXiv preprint arXiv:1912.10207}, 2019.

\bibitem{krizhevsky2010cifar}
A.~Krizhevsky and G.~Hinton, ``{Learning Multiple Layers of Features from Tiny
  Images},'' University of Toronto, Tech. Rep., 2009.

\bibitem{deng2009imagenet}
J.~Deng, W.~Dong, R.~Socher, L.-J. Li, K.~Li, and L.~Fei-Fei, ``Imagenet: A
  large-scale hierarchical image database,'' \emph{IEEE/CVF Int. Conf. on
  Computer Vision}, pp. 248--255, 2009.

\bibitem{lee2015deeply}
C.-Y. Lee, S.~Xie, P.~Gallagher, Z.~Zhang, and Z.~Tu, ``{Deeply-Supervised
  Nets},'' \emph{Int. Conf. on Artificial Intelligence and Statistics},
  vol.~38, pp. 562--570, 2015.

\bibitem{he2016deep}
K.~He, X.~Zhang, S.~Ren, and J.~Sun, ``{Deep Residual Learning for Image
  Recognition},'' in \emph{IEEE/CVPR Conf. on Computer Vision and Pattern
  Recognition}, 2016, pp. 770--778.

\bibitem{he2015delving}
------, ``{Delving Deep into Rectifiers: Surpassing Human-Level Performance on
  ImageNet Classification},'' in \emph{IEEE Int. Conf. on Computer Vision},
  2015, pp. 1026--1034.

\bibitem{zhang2018lq}
D.~Zhang, J.~Yang, D.~Ye, and G.~Hua, ``{LQ-Nets: Learned Quantization for
  Highly Accurate and Compact Deep Neural Networks},'' in \emph{Proceedings of
  the European conference on computer vision (ECCV)}, 2018, pp. 365--382.

\bibitem{jain2020trained}
S.~Jain, A.~Gural, M.~Wu, and C.~Dick, ``{Trained Quantization Thresholds for
  Accurate and Efficient Fixed-Point Inference of Deep Neural Networks},''
  \emph{Machine Learning and Systems}, vol.~2, pp. 112--128, 2020.

\end{thebibliography}
